%% file: root.tex
\documentclass[letterpaper, 10 pt, conference]{ieeeconf}  

\makeatletter
\let\NAT@parse\undefined
\makeatother

\IEEEoverridecommandlockouts                              

\usepackage{amsmath,amsfonts}
\usepackage{algorithmic}
\usepackage{algorithm}
\usepackage{amsmath}
\usepackage{array}
\usepackage{bm}
\usepackage[caption=false,font=footnotesize,labelfont=rm,textfont=rm]{subfig}  
\usepackage{textcomp}
\usepackage{stfloats}
\usepackage{url}
\usepackage{verbatim}
\usepackage{graphicx}
\usepackage{color}
\usepackage[colorlinks,linkcolor=blue,anchorcolor=blue,citecolor=blue]{hyperref}
\usepackage{siunitx}
\usepackage{multirow, booktabs}
\usepackage[numbers,sort&compress]{natbib}
\usepackage{threeparttable}

\usepackage{titlesec}
\titlespacing{\subsection}{0pt}{4.5pt}{2pt}

\newcommand{\MethodName}{CalibFormer}

\title{\LARGE \bf
\MethodName: A Transformer-based Automatic LiDAR-Camera Calibration Network}

\author{Yuxuan Xiao$^{1}$, Yao Li$^{1}$, Chengzhen Meng$^{1}$, Xingchen Li$^{1}$, Jianmin Ji$^{1}$ and Yanyong Zhang$^{2}$ 
\thanks{*The work is partially supported by the National Natural Science Foundation of China (No.62332016) and Anhui Province Development and Reform Commission 2021 New Energy and Intelligent Connected Vehicle Innovation Project.}
\thanks{$^{1}$Yuxuan Xiao, Yao Li, Chengzhen Meng, Xingchen Li, and Jianmin Ji are with School of Computer Science and Technology, University of Science and Technology of China, Hefei, China. (e-mail: \{xiaoyx, zkdly, czmeng, starlet\}@mail.ustc.edu.cn, jianmin@ustc.edu.cn)}
\thanks{$^{2}$Yanyong Zhang is the corresponding author, with School of Computer Science and Technology, University of Science and Technology of China, Hefei, China, and also with Institute of Artificial Intelligence, Hefei Comprehensive National Science Center, Hefei, China. (e-mail: yanyongz@ustc.edu.cn)}
}

\begin{document}

\maketitle
\thispagestyle{empty}
\pagestyle{empty}


\input{Chapters/Abstract.tex}
\input{Chapters/Introduction.tex}
\input{Chapters/RelatedWorks.tex}
\input{Chapters/Methods.tex}
\input{Chapters/Experiments.tex}
\input{Chapters/Conclusion.tex}

\footnotesize
\bibliographystyle{IEEEtranN}
\bibliography{refs}

\end{document}

%% file: Chapters/Abstract.tex
\begin{abstract}
The fusion of LiDARs and cameras has been increasingly adopted in autonomous driving for perception tasks. The performance of such fusion-based algorithms largely depends on the accuracy of sensor calibration, which is challenging due to the difficulty of identifying common features across different data modalities. Previously, many calibration methods involved specific targets and/or manual intervention, which has proven to be cumbersome and costly. Learning-based online calibration methods have been proposed, but their performance is barely satisfactory in most cases. These methods usually suffer from issues such as sparse feature maps, unreliable cross-modality association, inaccurate calibration parameter regression, etc. In this paper, to address these issues, we propose \MethodName, an end-to-end network for automatic LiDAR-camera calibration. We aggregate multiple layers of camera and LiDAR image features to achieve high-resolution representations. A multi-head correlation module is utilized to identify correlations between features more accurately. Lastly, we employ transformer architectures to estimate accurate calibration parameters from the correlation information. Our method achieved a mean translation error of $0.8751 \si{\cm}$ and a mean rotation error of $0.0562 \si{\degree}$ on the KITTI dataset, surpassing existing state-of-the-art methods and demonstrating strong robustness, accuracy, and generalization capabilities.
\end{abstract}

%% file: Chapters/Introduction.tex
\section{Introduction}

Nowadays, LiDARs and cameras have played a critical role in robotic systems such as autonomous vehicles. Cameras capture high-resolution images with detailed color and texture information, while LiDARs provide precise 3D point cloud representations of the environment. The fusion of LiDARs and cameras has been widely adopted in many tasks, such as 3D object detection~\cite{bai2022transfusion, chen2022autoalignv2, li2022ezfusion} and SLAM~\cite{shan2021lvisam, lin2022r3live}, achieving better performance than relying on a single modality. However, a crucial prerequisite for successful multi-sensor fusion is extrinsic calibration, which involves determining the 6-degree-of-freedom (6-DoF) transformation between the coordinate systems of the sensors.

The main challenge in calibration problems is to effectively identify and correlate common features across different data modalities. Many existing calibration methods~\cite{geiger2012automatic, gong20133d, kummerle2018automatic} rely on calibration targets, such as checkerboards or boards with specific patterns. They detect, extract, and match pairwise calibration targets between 2D images and 3D point clouds, which transforms the calibration problem into an optimization problem. These methods typically achieve satisfactory results with predefined external targets. However, calibration parameters can vary irregularly due to factors such as temperature changes or disturbances during vehicle movements.
Even a slight relative position offset between LiDAR and camera necessitates a recalibration to correct the extrinsic parameter drift. Consequently, timely target-based methods become prohibitively expensive, if at all possible, which renders targetless calibration approaches necessary.

Some works~\cite{pandey2012automatic, taylor2015motion, levinson2013automatic} have attempted to obtain the transformation matrix using targetless methods, but they require hand-crafted features and may not work well in different environments.
Recently, deep learning techniques have been widely used and demonstrated the superiority of automatic feature engineering.
For example, several methods~\cite{schneider2017regnet, shi2020calibrcnn, lv2021lccnet} use deep learning models to estimate the 6-DoF transformation between camera and LiDAR coordinates. Most deep learning-based methods utilize RGB images and depth images obtained through point cloud projection {\textemdash} often miscalibrated {\textemdash} as inputs. The processing pipeline primarily involves three steps: feature extraction, feature matching, and calibration parameter regression.
However, deep learning-based calibration methods face several important design issues. For example, the balance between calibration accuracy and computation efficiency is hard to achieve. It is also challenging to accurately correlate corresponding features from different modalities due to the disparate physical characteristics and operational principles of sensors, yielding data of varying dimensions, qualities, and types. Furthermore, not all correlated features contribute equally to calibration, making it imperative to extract feature correlations with higher contributions. 

In this paper, we present \MethodName, an automatic calibration method for LiDARs and cameras to address these issues. 
Firstly, calibration usually requires precise alignment between modalities, which in turn demands precise representations of sensor data contributions. Therefore, in the feature extraction phase, we upsample and aggregate multi-layer features~\cite{zhou2020tracking} from RGB and LiDAR data, obtaining fine-grained representations of the features.
In the matching phase, we apply a multi-head correlation module to calculate multi-dimensional correlation representations. In order to capture sufficient correlations from unaligned features, we treat feature correlation as a learnable implicit representation and employ a multi-head computation approach.
Subsequently, leveraging the computed correlation features, we employ a Swin Transformer~\cite{liu2022swin} encoder and a Transformer~\cite{vaswani2017attention} decoder to process the correlation feature effectively. Finally, a feed-forward network is utilized to regress the translation and rotation parameters separately.
In summary, our contributions can be summarized as follows: 

\begin{itemize}
    \item {\MethodName} is an end-to-end network designed for LiDAR-camera calibration. We extract fine-grained feature maps to achieve precise correlations and take a trade-off between computation and performance.
    \item We apply a multi-head correlation module to calculate correspondences between misaligned features across different dimensions. Then we utilize the transformer architecture to extract and leverage correlation features with higher contributions.
    \item The experiment results demonstrate that our approach achieves a translation error of $0.8751 \si{\cm}$ and a rotation error of $0.0562 \si{\degree}$, surpassing other deep learning-based methods and exhibiting robust generalization capabilities. Ablation experiments also validate the effectiveness of various modules. 
\end{itemize}

%% file: Chapters/RelatedWorks.tex
\section{Related Works}

Existing extrinsic calibration methods can be broadly categorized into target-based and targetless methods. The primary difference between them is the use of specific targets during the calibration process. Target-based methods require artificial calibration objects that offer explicit geometric features in both modalities. In contrast, targetless methods do not rely on any designated targets but instead utilize information from the surrounding environment.

\subsection{Target-based Methods}

\citet{zhang2004extrinsic} first proposed the target-based extrinsic calibration method using a checkerboard. The algorithm detects the grid corner to estimate its pose relative to the checkerboard in the camera image. It then calculates the extrinsic parameters between the laser scanner and the camera. The transformation is formulated into a nonlinear optimization problem.
\citet{geiger2012automatic} further applies a similar optimization problem to calibrate the camera and the 3D LiDAR. Other targets, such as trihedral~\cite{gong20133d} or spherical targets~\cite{kummerle2018automatic, toth2020automatic}, are also used for calibration besides the checkerboard.
\citet{beltran2022automatic} uses a rectangular plate with 4 holes as the target and affixes 4 ArUco tags~\cite{garrido2014automatic}. They detect and match the center points of these 4 holes and calculate the transformation between the sensors.
In general, these methods extract geometric features, such as points, edges, or planes, from images and point clouds according to the target and then match them. The calibration problem is then formulated into an optimization problem.
Despite being simple and effective, target-based methods are still time-consuming and laborious. Moreover, these methods cannot be used everywhere due to the limitations of the targets. 

\subsection{Targetless Methods}

\citet{li2023automatic} divided targetless calibration methods into four categories: information theory-based, feature-based, ego-motion-based, and learning-based. 

\citet{pandey2012automatic} proposed a mutual information-based algorithm that uses the correspondence between the intensity of point clouds and the grayscale of images.
\citet{taylor2015motion} utilize the ego-motion of sensors mounted on the moving vehicle to estimate extrinsic parameters.
\citet{levinson2013automatic} and \citet{yuan2021pixel} respectively extract depth-discontinuous and depth-continuous edge features and then match them to minimize the objective function.

Regnet~\cite{schneider2017regnet} was the first to adopt a deep learning approach. It extracts and matches features using a network before regressing the calibration parameters.
CalibNet~\cite{iyer2018calibnet} incorporates geometric information by introducing a 3D spatial transformer layer into the model.
CalibRCNN~\cite{shi2020calibrcnn} combines CNN with LSTM and adds pose constraints between consecutive frames to improve the calibration accuracy.
LCCNet~\cite{lv2021lccnet} utilizes the cost volume to compute the correlation between features from different sensors. 
Moreover, in sensor fusion perception tasks such as object detection~\cite{bai2022transfusion, li2022deepfusion, wan2022one}, the transformer's cross-attention mechanism is used to achieve soft association between different modalities. Nevertheless, the application of the transformer in calibration methods is rare.

In contrast to these methods, our approach employs a deep layer aggregation module to obtain high-resolution feature maps from the LiDAR and camera. Furthermore, we incorporate a multi-head correlation module to compute inter-sensor correlations in a more fine-grained manner. We also utilize a transformer architecture to process correlations and estimate calibration parameters effectively.

%% file: Chapters/Methods.tex
\begin{figure*}[tbp]
    \centering
    \setlength{\abovecaptionskip}{-5pt}
    \includegraphics[width=0.95\textwidth]{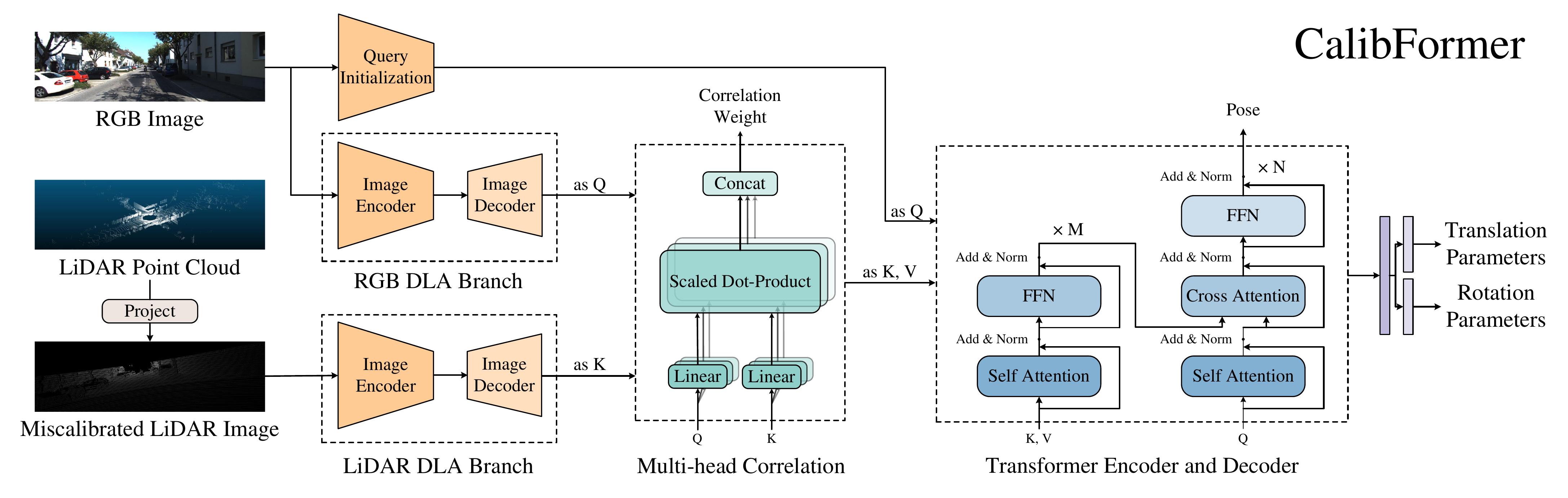}
    \caption{The overview of our proposed method for camera and LiDAR calibration. Firstly, we project the LiDAR point cloud onto the image plane, generating a miscalibrated LiDAR image using the initial extrinsic parameter $\mathbf{T}_{init}$ and the camera matrix $\mathbf{K}$. Our network takes both camera images and LiDAR images as inputs. After extracting fine-grained features, we employ a multi-head correlation module and a transformer architecture to obtain a 6-DoF transformation $\mathbf{T}_{pred}$ representing the deviation between the initial extrinsic parameter $\mathbf{T}_{init}$ and the accurate extrinsic parameter $\mathbf{T}_{LC}$.}
    \label{fig:overview}
    \vspace{-18pt}
\end{figure*}

\section{Methods}
Our network takes a pair of misaligned camera images and LiDAR point cloud as inputs and outputs the deviation $\mathbf{T}_{pred}$ of the initial calibration parameters $\mathbf{T}_{init}$ from the ground truth $\mathbf{T}_{LC}$. It first extracts fine-grained features from both modalities separately and calculates their correlations through a multi-head correlation module. Finally, the network regresses the deviations of the calibration parameters. The workflow of our proposed network is shown in Fig.~\ref{fig:overview}.

\subsection{Input Data Preprocessing}
\label{sec:preprocess}
We first project the LiDAR point cloud onto the image plane to obtain a sparse depth map. For each 3D point $p_{i}^{L}=[x_{i}\ y_{i}\ z_{i}]^T \in \mathbb{R}^3$ in the point cloud, given an initial extrinsic parameter $\mathbf{T}_{init}$ and camera intrinsic matrix $\mathbf{K}$, we can project it onto a 2D coordinate system on the image plane, denoted as pixel $p_{i}^{I}=[u_{i}\ v_{i}]^T \in \mathbb{R}^2$. The projection process can be expressed as:
\begin{equation} \small
    \begin{aligned}
        d_{i} \left [ \begin{matrix} u_{i} \\ v_{i} \\ 1 \end{matrix} \right ] &= \mathbf{K} \mathbf{T}_{init} \left [ \begin{matrix} x_{i} \\ y_{i} \\ z_{i} \\ 1 \end{matrix} \right ] = \mathbf{K} \left [ \begin{matrix} \mathbf{R}_{init} & \mathbf{t}_{init} \\ \mathbf{0} & 1 \end{matrix} \right ] \left [ \begin{matrix} x_{i} \\ y_{i} \\ z_{i} \\ 1 \end{matrix} \right],
    \end{aligned}
    \label{eq:projection}
\end{equation}
where $\hat{p}_{i}^{I}=[u_{i}\ v_{i}\ 1]^T$ and $\hat{p}_{i}^{L}=[x_{i}\ y_{i}\ z_{i}\ 1]^T$ represent the homogeneous coordinates of $p_{i}^{I}$ and $p_{i}^{L}$, respectively. $\mathbf{R}_{init}$ and $\mathbf{t}_{init}$ represent the initial rotation matrix and translation vector of extrinsic parameter $\mathbf{T}_{init}$. The depth of $p_{i}^{L}$ after projection onto the image plane is represented by $d_{i}$, which is used to construct the depth map $X_{dm}$ along with $p_{i}^{I}$.

To further utilize the LiDAR data, we also record the intensity $I_{i}$ of each LiDAR point $p_{i}^{L}$ during projection. Together with depth $d_{i}$, we can obtain a two-channel LiDAR image $X_{lidar} \in \mathbb{R}^{2 \times H \times W}$, where $H$ and $W$ represent the height and width of the image.

\subsection{Fine-grained Feature Extraction}
\begin{figure}[tbp]
    \centering
    \setlength{\abovecaptionskip}{-5pt}
    \includegraphics[width=0.45\textwidth]{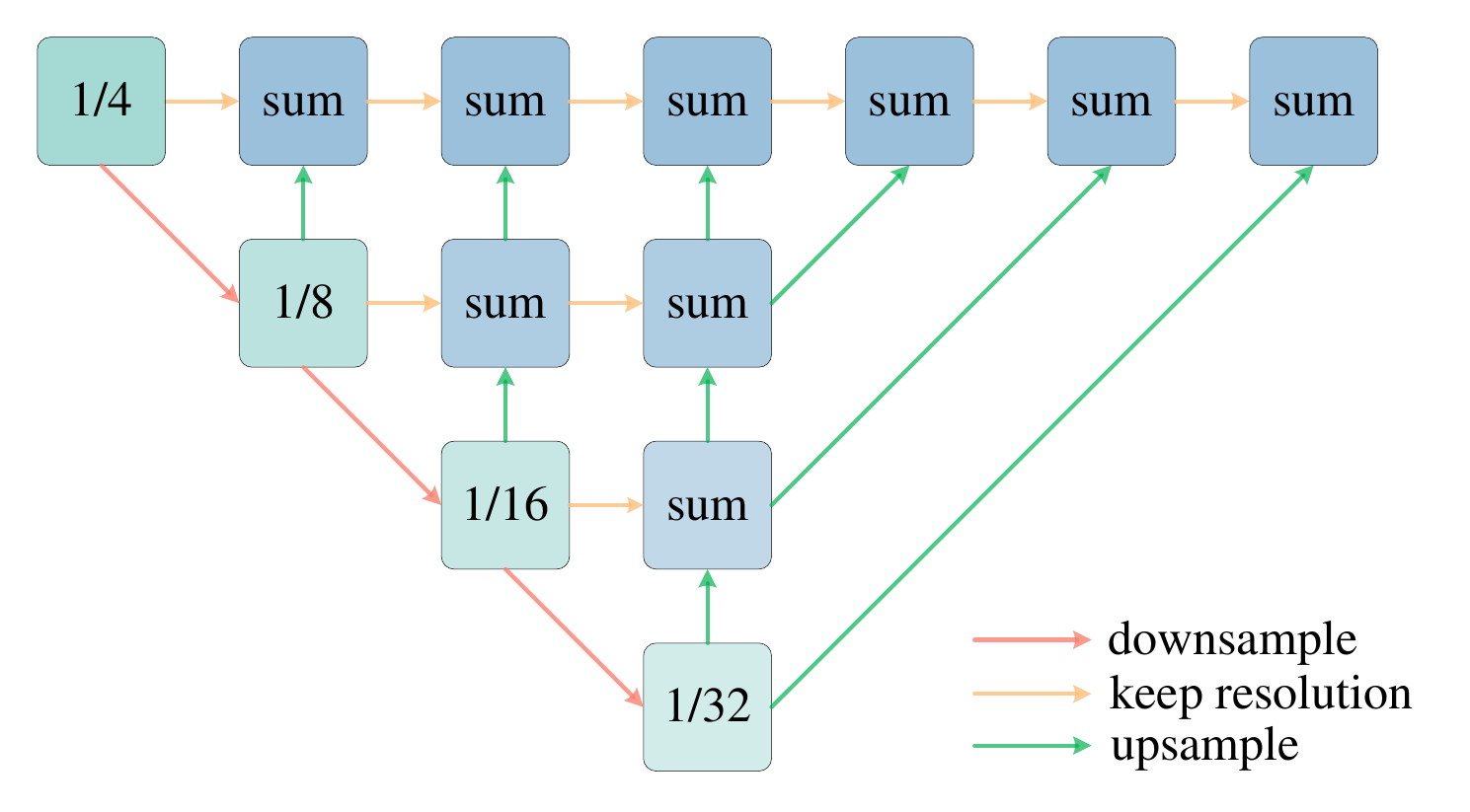}
    \caption{Overview of deep layer aggregation. After obtaining the features generated by the backbone at different layers, these features are respectively upsampled and aggregated to obtain a high-resolution feature map. }
    \label{fig:dla}
    \vspace{-18pt}
\end{figure}
The feature extraction network consists of two symmetric branches for extracting features from the camera image $X_{cam}$ and LiDAR image $X_{lidar}$, respectively. In the camera branch, we use a pre-trained ResNet-18~\cite{he2016deep} model to extract image features. Meanwhile, the LiDAR branch has a similar structure to the image branch, except that the number of input channels for the first convolutional layer is adjusted to two to accommodate the LiDAR image $X_{lidar}$.

We use an enhanced version of the Deep Layer Aggregation (DLA)~\cite{zhou2020tracking} to fuse multi-scale features with different receptive fields, as illustrated in Fig.~\ref{fig:dla}. Unlike the original DLA~\cite{yu2018deep}, it has more skip connections between feature maps of different scales. Additionally, to dynamically adjust the receptive field based on the image, the convolutional layers in the upsampling module of the DLA are replaced with deformable convolution layers~\cite{zhu2019deformable}. We apply this enhanced DLA on both the camera and LiDAR branches separately, resulting in high-resolution feature maps $F_{lidar}$ and $F_{cam}$, which are used for subsequent calibration.

We additionally employ a camera-guided query initialization module to extract features $F_{query}$ from camera images. This module consists of a ResNet-18 and a global average pooling layer, employed for generating the initial pose query.

\subsection{Multi-head Feature Matching}
After separately extracting fine-grained features from the input data, a feature matching module is used to calculate the correlation between the misaligned features of the two modalities caused by the noisy initial parameter $\mathbf{T}_{init}$ during LiDAR point projection in Section~\ref{sec:preprocess}. To compute the correlation between features in a fine-grained manner, we apply a multi-head correlation module.

Firstly, the LiDAR feature map $F_{lidar}$ and camera feature map $F_{cam}$ are flattened into queries $Q$ and keys $K$, both with a dimension of $d_{k}$.
Then, we compute the dot product between the query $Q$ and the key $K$.
After that, we divide the result by $\sqrt{d_{k}}$ to obtain the correlation weights. The computation process can be expressed as: 
\begin{equation} \small
    \mathrm{Correlation}(Q,K)=\frac{QK^T}{\sqrt{d_k}}.
\end{equation}
The correlation weights reflect the correlation between the two modalities.

To capture different correlations between queries and keys in multiple dimensions, we use $n$ sets of linear projections to transform the queries and keys independently. We compute the correlation weights separately for each set and then concatenate them to obtain the final output. The process is depicted below:
\begin{equation} \small
    \begin{aligned}
        \mathrm{MultiHead}(Q,K)&=\mathrm{Concat}(\mathrm{head}_{1},\dots,\mathrm{head}_{n}), \\
        \mathrm{head}_{i}&=\mathrm{Correlation}(QW_{i}^{Q},KW_{i}^{K}),
    \end{aligned}
\end{equation}
where $W_{i}^{Q}$ and $W_{i}^{K}$ are linear projection matrices.

Given that the initial deviation of extrinsic parameters falls within a certain range, the offset of the LiDAR and camera image features is also expected to be within a certain range. Specifically, for an image with a resolution of $1241 \times 376$ from the KITTI dataset, if the translation and rotation deviations of the calibration parameters are within $0.5 \si{\m}$ and $5 \si{\degree}$ respectively, the projected point cloud deviation is approximately within $100$ pixels. To leverage this characteristic, we apply a window-based approach, in which only the correlation weights within a particular window are utilized for calibration. Given a window size $d$, the correlation weights between the camera feature $F_{cam}(p_{i})$ and the LiDAR feature $F_{lidar}(p_{j})$ are only considered if $|p_{i}-p_{j}|_{\infty} \leq d$, where $p_i$ and $p_j$ are the 2D positions of camera feature map $F_{cam}$ and LiDAR feature map $F_{lidar}$, respectively. Therefore, we can obtain the correlation feature map $F_{corr} \in \mathbb{R}^{((2d+1)^2 \times n) \times H \times W}$, where $H$ and $W$ denote the height and width of the feature map, respectively, and $n$ denotes the number of heads.

\subsection{Transformer-based Parameter Regression}
We employ a transformer architecture to extract calibration parameters from the correlation feature map $F_{corr}$. For the correlation feature $F_{corr}$, we first increase its dimension to $d_{k}$ by densely connected convolutional layers~\cite{huang2017densely}. Subsequently, it is flattened and employed as input $Q$, $K$, and $V$ for the transformer encoder. Given the high resolution of the correlation feature map, we utilize a Swin Transformer encoder~\cite{liu2022swin} to mitigate computational complexity. We employ a Transformer decoder~\cite{vaswani2017attention} to estimate calibration parameters. The query $Q$ is derived from the initial pose query $F_{query}$, while the key $K$ and value $V$ are obtained from the encoder's output. The positions are embedded as position encoding using a multi-layer perceptron (MLP) and element-wise added to the key features. Finally, a feed-forward network (FFN) is utilized to regress the pose information, and then we obtain the translation parameters and rotation parameters separately. The translation parameter is represented by a $1 \times 3$ vector $\mathbf{t}_{pred}$, and the rotation parameter is represented by a $1 \times 4$ quaternion $\mathbf{q}_{pred}$.

\subsection{Loss Function}
To guide the network convergence, we employ the following loss function:
\begin{equation} \small
    L=\lambda_{t}L_{t}+\lambda_{r}L_{r}+\lambda_{p}L_{p},
\end{equation}
where $\lambda_{t}$, $\lambda_{r}$ and $\lambda_{p}$ represent the weights for translation loss, rotation loss, and point cloud distance loss, respectively.

For the translation part, we used the smoothed L1 loss to represent the error:
\begin{equation} \small
    L_{t}=\mathrm{Smooth L1}(\mathbf{t}_{gt}-\mathbf{t}_{pred}).
\end{equation}

Regarding the rotation component, to avoid the double-covering issue of quaternions, we use the angular distance to measure the difference between the quaternions. For a unit quaternion $\mathbf{q}$, using $\Re(\mathbf{q})$ and $\Im(\mathbf{q})$ to respectively represent its real part and imaginary part, the rotation loss is defined as follows:
\begin{equation} \small
    L_{r}=D_{a}(\mathbf{q}_{gt}, \mathbf{q}_{pred})=2 \arctan(\frac{\Vert\Im(\mathbf{q}_{gt} \times \mathbf{q}_{pred}^{-1})\Vert}{\Re(\mathbf{q}_{gt} \times \mathbf{q}_{pred}^{-1})}).
\end{equation}

The point cloud distance loss describes the distance between corresponding points in the point clouds before and after the extrinsic transformation:
\begin{equation} \small
    L_{p}=\frac{1}{N} \sum_{i=1}^{N} \Vert \mathbf{T}_{gt}^{-1}\mathbf{T}_{pred}\mathbf{p}_{i}^{L}-\mathbf{p}_{i}^{L} \Vert,
\end{equation}
where $N$ donates the number of points in this point cloud.

\subsection{Calibration Inference}
We denote the rotation component as a rotation matrix $\mathbf{R}_{pred}$ converted from the network prediction quaternion value $\mathbf{q}_{pred}$. By concatenating the rotation matrix $\mathbf{R}_{pred}$ with the translation vector $\mathbf{t}_{pred}$, we obtain the predicted error:
\begin{equation} \small
    \mathbf{T}_{pred} = \left [ \begin{matrix} \mathbf{R}_{pred} & \mathbf{t}_{pred} \\ \mathbf{0} & 1 \end{matrix} \right ].
\end{equation}

Given the initial extrinsic parameter $\mathbf{T}_{init}$ and the predicted error $\mathbf{T}_{pred}$, the calibration parameters can be obtained as follows:
\begin{equation} \small
    \widehat{\mathbf{T}}_{LC}=\mathbf{T}_{pred}^{-1}\mathbf{T}_{init}.
\end{equation}

%% file: Chapters/Experiments.tex
\section{Experiments}

\subsection{Dataset Preparation}

We evaluate our method on KITTI odometry dataset~\cite{geiger2012are}, which consists of 22 sequences from different scenes. We use sequences 01 to 21 as the training and validation set, and sequence 00 as the test set. To acquire a sufficient amount of training data, we introduce uniformly distributed random deviations $\Delta \mathbf{T}$ within a certain range to the extrinsic parameters of the data.
We obtain the initial extrinsic parameter $\mathbf{T}_{init}=\Delta \mathbf{T} \times \mathbf{T}_{LC}$, and use it along with the point cloud and image as input. The ground truth is denoted as $\mathbf{T}_{gt}=\Delta \mathbf{T}$.
A significant amount of training data can be obtained by randomly generating deviations.

\subsection{Implementing Details}

\vspace{0pt}\noindent\textbf{Training details:}
The original resolution of the image is padded to $1280 \times 384$ and subsequently resized to $512 \times 256$ as the input for the model.
The upsampling rate for feature maps is set to $4$. The window size in the multi-head correlation module is set to $4$. Furthermore, the Swin Transformer encoder is configured with $2$ layers, while the Transformer decoder comprises $6$ layers.
We train the network using the Adam optimizer~\cite{kingma2017adam} with a learning rate of $5e^{-4}$ for $500$ epochs and a batch size of $256$.
Latency is measured on an NVIDIA RTX 3060 GPU.

\vspace{2pt}\noindent\textbf{Evaluation Metrics:}
The calibration results are evaluated by translation and rotation parameters. For the translation component, we separately record mean absolute errors in the X, Y, and Z directions.
Regarding the rotation component, although it can be represented using quaternions, we opt to convert it into Euler angles to provide a more intuitive assessment of rotation error. Similarly, we separately record mean absolute errors of roll, pitch, and yaw.

\begin{table*}[tbp]
    \centering
    \begin{threeparttable}
        \caption{Comparison of different methods on the KITTI odometry dataset. The upper half is the result of a deviation of $(\pm 0.25 \si{\m}, \pm 10 \si{\degree})$ and the lower half is the result of a deviation of $(\pm 0.5 \si{\m}, \pm 5 \si{\degree})$.}
        \label{tab:results}
        \begin{tabular}{cccccccccc}
            \toprule
            \multirow{2}[2]{*}{Deviation} & \multirow{2}[2]{*}{Method} & \multicolumn{4}{c}{Translation(\si{\cm}) $\downarrow$} & \multicolumn{4}{c}{Rotation(\si{\degree}) $\downarrow$} \\
            \cmidrule(lr){3-6} \cmidrule(lr){7-10}
             & & Mean & X & Y & Z & Mean & Roll & Pitch & Yaw \\
            \midrule
            \multirow{4}{*}{$(\pm 0.25 \si{\m}, \pm 10 \si{\degree})$\tnote{1}} & CalibRCNN~\cite{shi2020calibrcnn} & $5.3$ & $6.2$ & $4.3$ & $5.4$ & $0.428$ & $0.199$ & $0.64$ & $0.446$ \\
            & CALNet~\cite{shang2022calnet} & $3.03$ & $3.65$ & $1.63$ & $3.80$ & $0.20$ & $0.10$ & $0.38$ & $0.12$ \\
            & PSNet~\cite{wu2022psnet} & $3.1$ & $3.8$ & $2.8$ & $2.6$ & $0.15$ & $\mathbf{0.06}$ & $0.26$ & $0.12$ \\
            & Ours & $\mathbf{1.1877}$ & $\mathbf{1.1006}$ & $\mathbf{0.9015}$ & $\mathbf{1.5611}$ & $\mathbf{0.1406}$ & $0.0764$ & $\mathbf{0.2588}$ & $\mathbf{0.0865}$ \\
            \midrule
            \multirow{3}{*}{$(\pm 0.5 \si{\m}, \pm 5 \si{\degree})$\tnote{2}} & LCCNet~\cite{lv2021lccnet} & $1.6737$ & $1.6695$ & $1.3192$ & $2.0325$ & $0.1576$ & $0.0556$ & $0.3080$ & $0.1091$ \\
            & CalibDepth~\cite{zhu2023calibdepth} & $0.9188$ & $0.9971$ & $\mathbf{0.5655}$ & $\mathbf{1.1938}$ & $0.1633$ & $0.0411$ & $\mathbf{0.0605}$ & $0.3883$ \\
            & Ours & $\mathbf{0.8751}$ & $\mathbf{0.7594}$ & $0.6027$ & $1.2633$ & $\mathbf{0.0562}$ & $\mathbf{0.0249}$ & $0.1041$ & $\mathbf{0.0395}$ \\
            \bottomrule
        \end{tabular}
        \begin{tablenotes}
            \footnotesize
            \item[1] Following CalibRCNN, we evaluate the method's performance under a deviation of $(\pm 0.25 \si{\m}, \pm 10 \si{\degree})$.
            \item[2] Following LCCNet, we evaluate the method's performance under a deviation of $(\pm 0.5 \si{\m}, \pm 5 \si{\degree})$.
        \end{tablenotes}
    \end{threeparttable}
    \vspace{-3pt}
\end{table*}

\subsection{Quantitative Results}

To compare performance with different methods, we follow their experimental setups, introducing two sets of initial deviations, $(\pm 0.5 \si{\m}, \pm 5 \si{\degree})$ and $(\pm 0.25 \si{\m}, \pm 10 \si{\degree})$, and evaluate the network's performance for each.
The results, as presented in Table~\ref{tab:results}, indicate that for both initial deviation settings, our proposed method outperforms other approaches.

\vspace{2pt}\noindent\textbf{Initial deviation $(\pm 0.25 \si{\m}, \pm 10 \si{\degree})$:} The mean translation error of our method is $1.1877 \si{\cm}$, and the mean rotation error is $0.1406 \si{\degree}$. Compared to PSNet~\cite{wu2022psnet}, our network shows a $61.7\%$ improvement in the translation performance and a $6.3\%$ improvement in the rotation performance.

\vspace{2pt}\noindent\textbf{Initial deviation $(\pm 0.5 \si{\m}, \pm 5 \si{\degree})$:} Our method exhibits a mean translation error of $0.8751 \si{\cm}$ and a mean rotation error of $0.0562 \si{\degree}$. Our network outperforms CalibDepth~\cite{zhu2023calibdepth} by $4.7\%$ in terms of translation performance and by $65.6\%$ in terms of rotation performance.

\begin{table*}[tbp]
    \centering
    \begin{threeparttable}
        \caption{Ablation experiments on KITTI odometry datasets.}
        \label{tab:ablation}
        \begin{tabular}{cccccccccc}
            \toprule
            \multirow{2}[2]{*}{Network Architecture} & \multicolumn{4}{c}{Translation(\si{\cm}) $\downarrow$} & \multicolumn{4}{c}{Rotation(\si{\degree}) $\downarrow$} & \multirow{2}[2]{*}{Latency(\si{\ms}) $\downarrow$} \\
            \cmidrule(lr){2-5} \cmidrule(lr){6-9}
             & Mean & X & Y & Z & Mean & Roll & Pitch & Yaw \\
            \midrule
            w/ all modules & $\mathbf{0.8751}$ & $\mathbf{0.7594}$ & $\mathbf{0.6027}$ & $\mathbf{1.2633}$ & $\mathbf{0.0562}$ & $\mathbf{0.0249}$ & $\mathbf{0.1041}$ & $\mathbf{0.0395}$ & $27.79$\\
            w/o multi-head correlation & $1.0970$ & $0.9300$ & $0.9408$ & $1.4203$ & $0.0691$ & $0.0332$ & $0.1245$ & $0.0496$ & $26.21$ \\
            w/o swin transformer encoder & $1.0809$ & $0.7727$ & $0.8980$ & $1.5720$ & $0.0742$ & $0.0383$ & $0.1375$ & $0.0414$ & $26.35$ \\
            w/o transformer architecture & $1.2366$ & $1.0320$ & $0.9645$ & $1.7133$ & $0.0750$ & $0.0426$ & $0.1258$ & $0.0566$ & $20.79$ \\
            w/o upsampling & $1.1617$ & $0.8904$ & $0.7955$ & $1.7991$ & $0.0985$ & $0.0350$ & $0.2054$ & $0.0550$ & $\mathbf{13.15}$ \\
            \bottomrule
        \end{tabular}
    \end{threeparttable}
    \vspace{-3pt}
\end{table*}

\begin{table*}[tbp]
    \centering
    \begin{threeparttable}
        \caption{Comparison of different upsampling rates.}
        \label{tab:upsample}
        \begin{tabular}{cccccccccc}
            \toprule
            \multirow{2}[2]{*}{Upsampling Rate} & \multicolumn{4}{c}{Translation(\si{\cm}) $\downarrow$} & \multicolumn{4}{c}{Rotation(\si{\degree}) $\downarrow$} & \multirow{2}[2]{*}{Latency(\si{\ms}) $\downarrow$} \\
            \cmidrule(lr){2-5} \cmidrule(lr){6-9}
             & Mean & X & Y & Z & Mean & Roll & Pitch & Yaw & \\
            \midrule
            $1 \times$ & $1.1617$ & $0.8904$ & $0.7955$ & $1.7991$ & $0.0985$ & $0.0350$ & $0.2054$ & $0.0550$ & $\mathbf{13.15}$ \\
            $2 \times$ & $1.0925$ & $0.8435$ & $0.7813$ & $1.6526$ & $0.0858$ & $0.0284$ & $0.1781$ & $0.0509$ & $19.66$ \\
            $4 \times$\tnote{*} & $0.8751$ & $0.7594$ & $\mathbf{0.6027}$ & $1.2633$ & $\mathbf{0.0562}$ & $0.0249$ & $\mathbf{0.1041}$ & $\mathbf{0.0395}$ & $27.79$ \\
            $8 \times$ & $\mathbf{0.8409}$ & $\mathbf{0.6839}$ & $0.6449$ & $\mathbf{1.1939}$ & $0.0576$ & $\mathbf{0.0242}$ & $0.1050$ & $0.0437$ & $39.32$ \\
            \bottomrule
        \end{tabular}
        \begin{tablenotes}
            \footnotesize
            \item[*] denotes the configuration used in our network.
        \end{tablenotes}
    \end{threeparttable}
    \vspace{-15pt}
\end{table*}

\subsection{Ablation Studies}

In this section, we compare the impact of several network architectures on calibration performance.
We perform a set of ablation experiments to show the effect of each component. The performance is evaluated under a deviation of $(\pm 0.5 \si{\m}, \pm 5 \si{\degree})$. The results of the ablation experiment are shown in Table~\ref{tab:ablation}. 

\vspace{2pt}\noindent\textbf{Multi-head Correlation:}
To investigate the influence of correlation computation methods on calibration results, we compare the multi-head correlation module with direct inner product computation. As shown in Table~\ref{tab:ablation}, the multi-head correlation module yields a $20.2\%$ improvement in translation performance and a $18.7\%$ improvement in rotation performance, indicating its ability to better correlate corresponding features. Our method achieved greater performance gains with only a minor increase in computational overhead.

\vspace{2pt}\noindent\textbf{Transformer Architecture:}
We investigate the impact of using a transformer architecture to process correlation features on calibration performance. As a comparison, we substitute a fully connected layer for the transformer architecture. As shown in Table~\ref{tab:ablation}, our approach yields a moderate increase in latency, resulting in a $29.2\%$ improvement in translation performance and a $25.1\%$ improvement in rotation performance. Furthermore, if only the Transformer decoder is utilized without the Swin Transformer encoder, the calibration results deteriorate. This outcome underscores the contribution of both the transformer encoder and decoder in locating and leveraging more valuable correlation features.

\vspace{2pt}\noindent\textbf{Upsampling:}
We compare the impact of aggregating the features of different layers, i.e., different upsampling rates. As shown in Table~\ref{tab:upsample}, we observe that using a higher upsampling rate leads to better results. This is because a higher-resolution output feature map allows for a more detailed representation of the input data, potentially enabling the network to capture more fine-grained features and patterns. However, it is important to consider that using a higher upsampling rate also increases the computational complexity and memory requirements of the network, involving there may be a trade-off between performance and efficiency. As the upsampling rate increases to $8$, the performance improvement becomes marginal compared to that of $4$, while the computational overhead is much higher. Hence, a $4 \times$ upsampling rate is a more appropriate choice.

\subsection{Qualitative Results}
\begin{figure*}[tbp]
    \centering
    \subfloat[]{
        \begin{minipage}[b]{0.3\textwidth}
            \includegraphics[width=\textwidth]{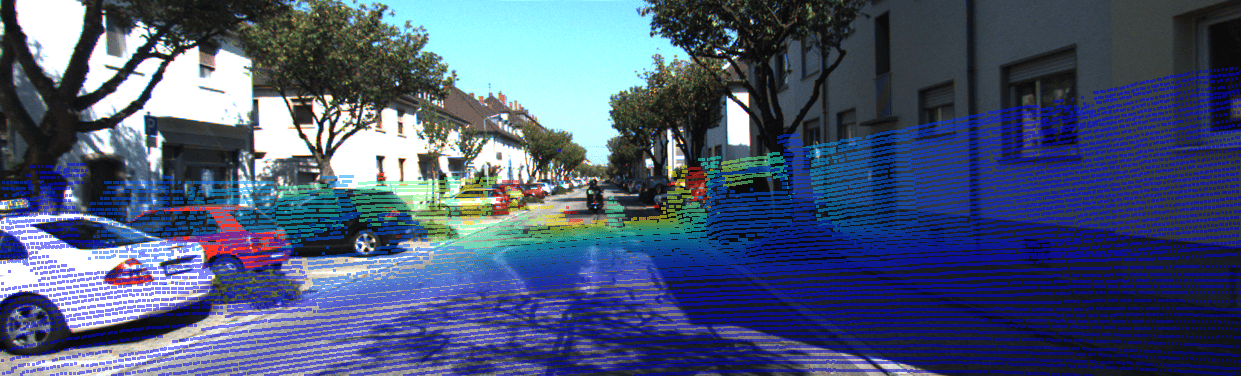}
            \includegraphics[width=\textwidth]{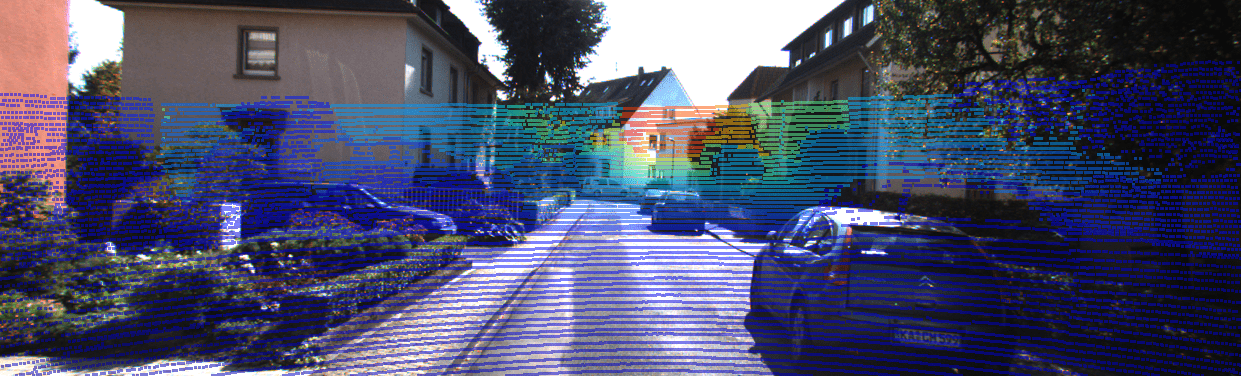}
            \includegraphics[width=\textwidth]{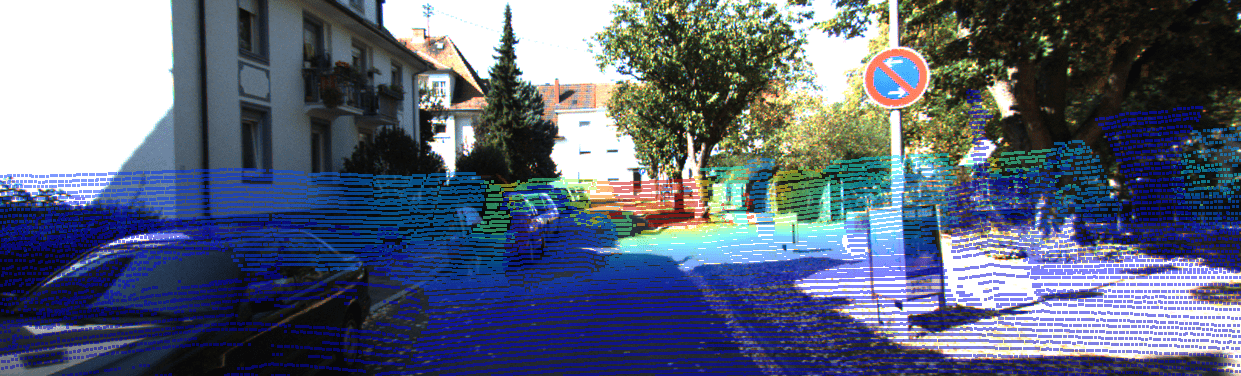}
        \end{minipage}
        \label{fig:results_input}
    }
    \subfloat[]{
        \begin{minipage}[b]{0.3\textwidth}
            \includegraphics[width=\textwidth]{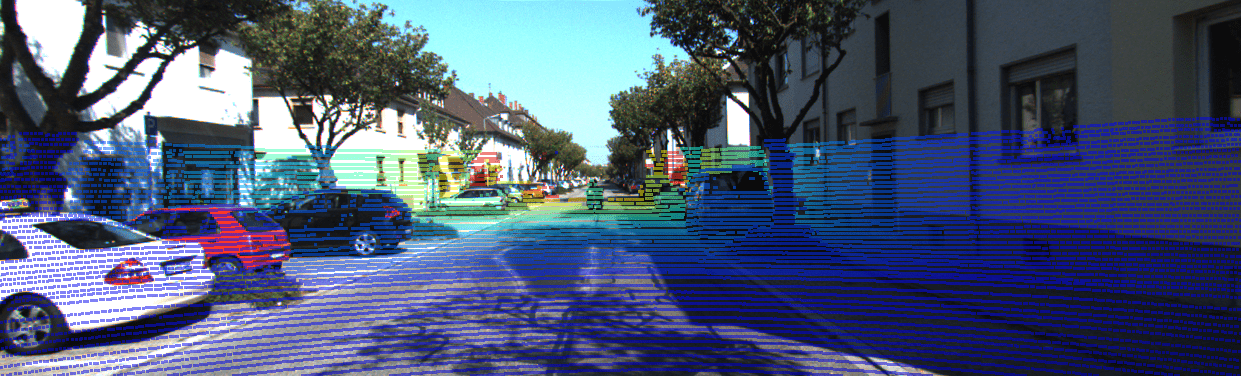}
            \includegraphics[width=\textwidth]{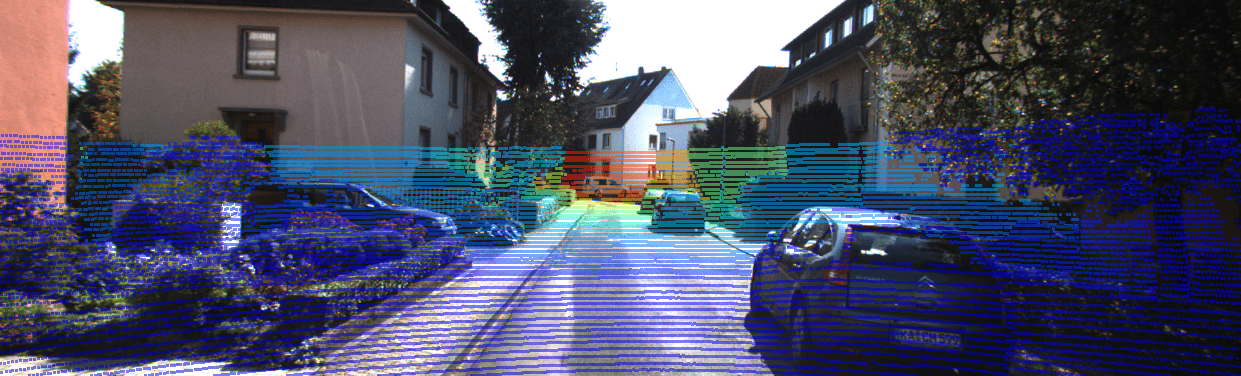}
            \includegraphics[width=\textwidth]{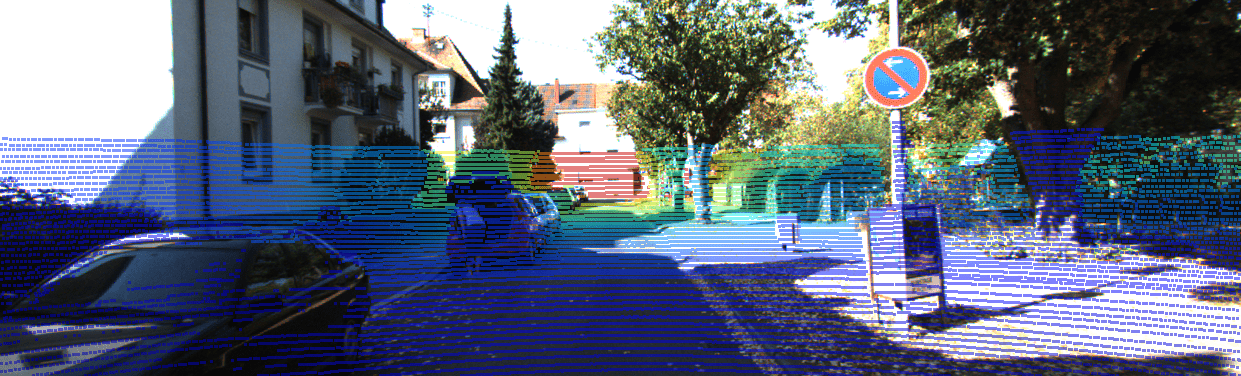}
        \end{minipage}
        \label{fig:results_pred}
    }
    \subfloat[]{
        \begin{minipage}[b]{0.3\textwidth}
            \includegraphics[width=\textwidth]{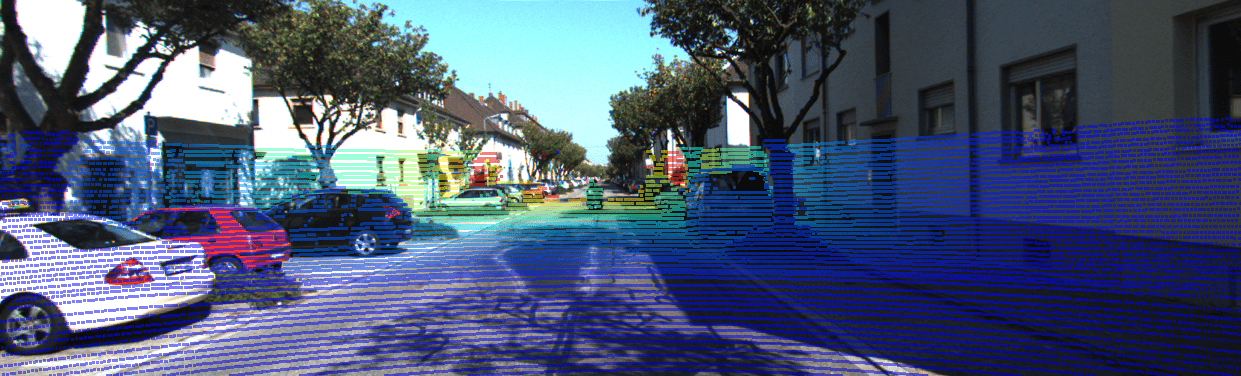}
            \includegraphics[width=\textwidth]{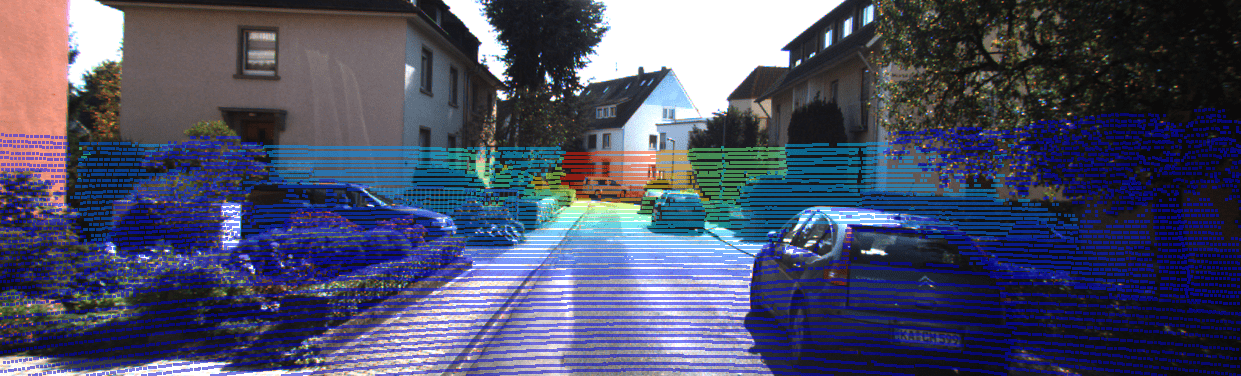}
            \includegraphics[width=\textwidth]{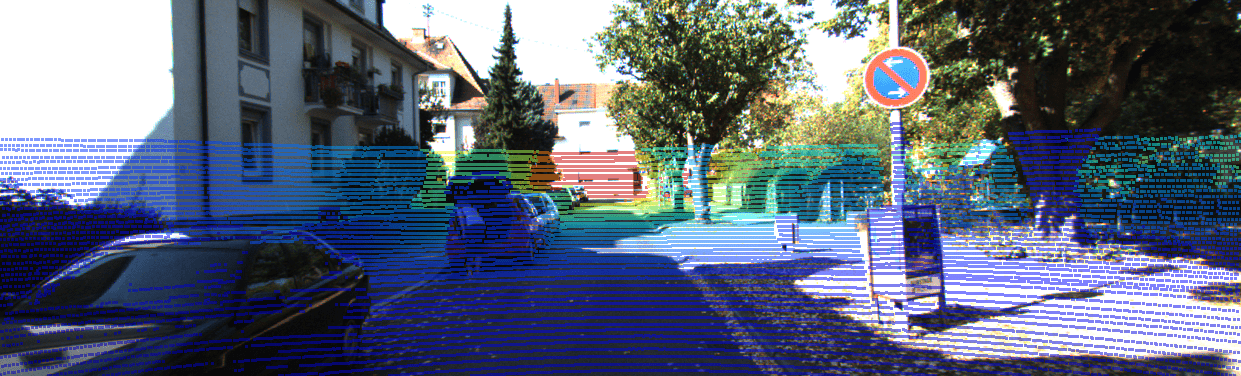}
        \end{minipage}
        \label{fig:results_gt}
    }
    \caption{Examples of calibration results for different scenes on the KITTI dataset. (a) represents the projection of miscalibrated point clouds onto the image plane. (b) shows the projection result of the point cloud using the network's predicted extrinsic parameters, and (c) represents the corresponding ground truth result.}
    \label{fig:results}
    \vspace{-10pt}
\end{figure*}

Our network is capable of achieving accurate calibration results under varying scenes and different initial miscalibrations. Fig.~\ref{fig:results} illustrates some of the visualized calibration results. We observe that even if initial extrinsic parameters exhibit errors in the direction of all six axes, our network is still able to align images and point clouds and generate results that are close to the ground truth.

\subsection{Generalization Validation}

\begin{table*}[tbp]
    \centering
    \begin{threeparttable}
        \caption{Calibration results on unfamiliar datasets and comparison with other methods.}
        \label{tab:generalization}
        \begin{tabular}{ccccccccc}
            \toprule
            \multirow{2}[2]{*}{Method} & \multicolumn{4}{c}{Translation(\si{\cm}) $\downarrow$} & \multicolumn{4}{c}{Rotation(\si{\degree}) $\downarrow$} \\
            \cmidrule(lr){2-5} \cmidrule(lr){6-9}
             & Mean & X & Y & Z & Mean & Roll & Pitch & Yaw \\
            \midrule
            LCCNet~\cite{lv2021lccnet} & $2.4896$ & $1.9233$ & $2.4894$ & $3.0561$ & $0.1937$ & $0.0868$ & $0.3647$ & $0.1297$ \\
            CalibDepth~\cite{zhu2023calibdepth} & $2.5413$ & $3.4261$ & $1.7634$ & $2.4345$ & $0.2818$ & $0.0689$ & $\mathbf{0.1546}$ & $0.6220$ \\
            Ours & $\mathbf{1.1037}$ & $\mathbf{1.0382}$ & $\mathbf{0.7043}$ & $\mathbf{1.5685}$ & $\mathbf{0.0805}$ & $\mathbf{0.0380}$ & $0.1556$ & $\mathbf{0.0479}$ \\
            \bottomrule
        \end{tabular}
    \end{threeparttable}
    \vspace{-15pt}
\end{table*}

The KITTI odometry dataset primarily consists of the ``2011\_09\_30'' and ``2011\_10\_03'' sequences from the KITTI raw dataset, along with a small part of the ``2011\_09\_26'' sequence. To evaluate our model's generalization capability, we train the network using the KITTI odometry dataset and subsequently evaluate its performance on the ``2011\_09\_26'' sequence of KITTI raw dataset, which includes unfamiliar scenes. Similarly, the initial deviation is set to $(\pm 0.5 \si{\m}, \pm 5 \si{\degree})$. The results shown in Table~\ref{tab:generalization} indicate that due to variations in the scene, the performance of our network on the KITTI raw dataset is worse compared to the results on the KITTI odometry dataset. Nevertheless, it still achieves a mean translation error of $1.1037 \si{\cm}$ and a mean rotation error of $0.0805 \si{\degree}$. In comparison to LCCNet~\cite{lv2021lccnet}, our method demonstrates a $55.7\%$ improvement in translation performance and a $58.4\%$ improvement in rotation performance. The test results on the KITTI raw dataset underscore the robustness of our network, demonstrating consistent and strong performance across diverse scenes.

%% file: Chapters/Conclusion.tex
\section{Conclusion}

In this paper, we propose an end-to-end calibration network for estimating the 6-DoF rigid body transformation between the LiDAR and the camera, which are important sensor combinations in autonomous driving systems for perception. 
Our network consists of three main parts: feature extraction module, feature matching module, and transformer regression module. The fine-grained feature extraction module employs DLA to aggregate multi-layer features to get a high-resolution feature representation. In the multi-head feature matching module, we use a multi-head correlation module to calculate the correlation between two modalities in a fine-grained manner. In the transformer-based parameter regression module, we employ both Swin Transformer and Transformer for encoding and decoding, resulting in accurate translation and rotation parameters. 
In order to solve the problem of insufficient training samples, we introduce random deviations to the extrinsic parameters to augment the training data. Our network can achieve an absolute error of $0.8751 \si{\cm}$ in translation and $0.0562 \si{\degree}$ in rotation, with initial miscalibrations up to $\pm 0.5 \si{\m}$ in translation and $\pm 5 \si{\degree}$ in rotation, outperforming other state-of-the-art methods.
Despite the increase in computational cost, the latency remains within a reasonable range.